\def\year{2018}\relax
\documentclass[letterpaper]{article} %DO NOT CHANGE THIS
\usepackage{aaai18}  %Required
\usepackage{times}  %Required
\usepackage{helvet}  %Required
\usepackage{courier}  %Required
\usepackage{url}  %Required
\usepackage{graphicx}  %Required
\usepackage{amssymb}
\usepackage{amsmath}
\begin{document}
\title{A Multilayer Convolutional Encoder-Decoder Neural Network \\for Grammatical Error Correction}
\author{Shamil Chollampatt$^{1}$ \and Hwee Tou Ng{$^{1,2}$}\\
    $^{1}$NUS Graduate School for Integrative Sciences and Engineering \\
 	$^{2}$Department of Computer Science \\
    National University of Singapore\\
   {shamil@u.nus.edu}, {nght@comp.nus.edu.sg}
 }
 
 \maketitle

%%%%%%%%%%%%%%%%
%% Abstract
%%%%%%%%%%%%%%%%
\begin{abstract}
We improve automatic correction of grammatical, orthographic, and collocation errors in text using a multilayer convolutional encoder-decoder neural network. The network is initialized with embeddings that make use of character \textit{N}-gram information to better suit this task. When evaluated on common benchmark test data sets (CoNLL-2014 and JFLEG), our model substantially outperforms all prior neural approaches on this task as well as strong statistical machine translation-based systems with neural and task-specific features trained on the same data. Our analysis shows the superiority of convolutional neural networks over recurrent neural networks such as long short-term memory (LSTM) networks in capturing the local context via attention, and thereby improving the coverage in correcting grammatical errors. By ensembling multiple models, and incorporating an \textit{N}-gram language model and edit features via rescoring, our novel method becomes the first neural approach to outperform the current state-of-the-art statistical machine translation-based approach, both in terms of grammaticality and fluency.
\end{abstract}

%%%%%%%%%%%%%%%%
%% Introduction
%%%%%%%%%%%%%%%%
\section{Introduction}
With the increasing number of non-native learners and writers of the English language around the globe, the necessity to improve authoring tools such as error correction systems is increasing. Grammatical error correction (GEC) is a well-established natural language processing (NLP) task that deals with building systems for automatically correcting errors in written text, particularly in non-native written text. The errors that a GEC system attempts to correct are not limited to grammatical errors, but also include spelling and collocation errors.

GEC in English has gained much attention within the NLP community recently. The phrase-based statistical machine translation (SMT) approach has emerged as the state-of-the-art approach for this task \cite{chollampatt2017connectingdots,junczysdowmunt2016phrase}, in which GEC is treated as a translation task from the language of ``bad'' English to the language of ``good'' English. The translation model is learned using parallel error-corrected corpora (source text that contains errors and their corresponding corrected target text). Although neural network (NN) models have been used as features to improve the generalization of the SMT approach \cite{chollampatt2016neural}, SMT still suffers from limitations in accessing the global source and target context effectively. The treatment of words and phrases as discrete entities during decoding also limits its generalization capabilities. To this end, several neural encoder-decoder approaches were proposed for this task \cite{xie2016neural,zheng2016grammatical,ji2017nested,schmaltz2017adapting}. However, their performance still falls substantially behind state-of-the-art SMT approaches.

All prior neural approaches for GEC relied on using recurrent neural networks (RNNs). In contrast to previous neural approaches, our neural approach to GEC is based on a fully convolutional encoder-decoder architecture with multiple layers of convolutions and attention \cite{gehring2017convolutional}. Our analysis shows that convolutional neural networks (CNNs) can capture local context more effectively than RNNs as the convolution operations are performed over smaller windows of word sequences. Most grammatical errors are often localized and dependent only on the nearby words. Wider contexts and interaction between distant words can also be captured by a multilayer hierarchical structure of convolutions and an attention mechanism that weights the source words based on their relevance in predicting the target word. Moreover, only a fixed number of non-linearities are performed on the input irrespective of the input length whereas in RNNs, the number of non-linearities is proportional to the length of the input, diminishing the effects of distant words. 

We further improve the performance by ensembling multiple models. Contrary to prior neural approaches, we use a simpler pre-processing method to alleviate the unknown word problem \cite{sennrich2016neural}. Rare words are split into multiple frequent sub-words using a byte pair encoding (BPE) algorithm. One of the major weaknesses of prior neural approaches is that they do not incorporate task-specific features nor utilize large native English corpora to good effect. We use such English corpora in our encoder-decoder model to pre-train the word vectors to be used for initializing the embeddings in the encoder and decoder. We also train an \textit{N}-gram language model to be used as a feature along with edit operation count features in rescoring to produce an overall better output.

To summarize, this paper makes the following contributions: (1) We successfully employ a convolutional encoder-decoder model trained on BPE tokens as our primary model to achieve state-of-the-art performance for GEC. Ours is the \emph{first} work to use fully convolutional neural networks for end-to-end GEC. (2) We exploit larger English corpora to pre-train word embeddings and to train an \textit{N}-gram language model to be used as a feature in a rescorer that is trained to optimize the target metric using minimum error rate training \cite{och2003mert}. (3) We conduct a comparison of attention mechanisms in typical recurrent architectures and our models, and perform error type performance analysis to identify the strengths of our approach over the current state-of-the-art SMT approach.

%%%%%%%%%%%%%%%%
%% Related Word
%%%%%%%%%%%%%%%%
\section{Related Work}

GEC gained much attention within the NLP community after the CoNLL-2014 shared task \cite{ng2014conll2014} was organized. The shared task dealt with the correction of all grammatical errors in English essays. Since then, the test set for the shared task has been used to benchmark GEC systems. Statistical machine translation has emerged as the state-of-the-art approach \cite{chollampatt2017connectingdots} due to its ability to correct various types of errors and complex error patterns, whereas previous approaches relied on building error type-specific classifiers \cite{dahlmeier-ng-ng:2012:BEA,rozovskaya2014illinois}. The SMT framework largely benefits from its ability to incorporate large error-corrected parallel corpora like the publicly available Lang-8 corpus \cite{mizumoto2011mining}, additional English corpora for training robust language models (LMs), task-specific features \cite{junczysdowmunt2016phrase}, and neural models \cite{chollampatt2016neural}. However, SMT-based systems suffer from limited generalization capabilities compared to neural approaches and are unable to access longer source and target contexts effectively. To address these issues, several neural encoder-decoder approaches relying on RNNs were proposed for GEC.

\subsection{Neural Encoder-Decoder Approaches to GEC}

\citeauthor{zheng2016grammatical}~\shortcite{zheng2016grammatical} first applied a popular neural machine translation model, \textit{RNNSearch} \cite{bahdanau2015neural}, consisting of a bidirectional RNN encoder and an attention-based RNN decoder. They additionally made use of an unsupervised word alignment model and a word-level statistical translation model to replace unknown words in the output. However, they trained their systems on l.9M sentence pairs from the professionally annotated, \textit{non-public} Cambridge Learner Corpus (CLC), making their models hard to replicate and compare with. 

\citeauthor{xie2016neural}~\shortcite{xie2016neural} proposed the use of a character-level recurrent encoder-decoder network for GEC. They trained their models on the publicly available NUCLE \cite{dahlmeier2013building} and Lang-8 corpora, along with synthesized  examples for frequent error types. They also incorporated an \textit{N}-gram LM trained on a small subset of the Common Crawl corpus (2.2B \textit{N}-grams) during decoding to achieve an F\textsubscript{0.5} score of 39.97 on the CoNLL-2014 test set. They further used a supervised edit classifier trained on character and word-level edit operation and pre-trained word embedding features to remove spurious edits and improve the F\textsubscript{0.5} score to 40.56.

\citeauthor{ji2017nested}~\shortcite{ji2017nested} proposed a hybrid word-character model based on the hybrid machine translation model of \cite{luong2016achieving}, by adding nested levels of attention at the word and character level. Similar to \cite{zheng2016grammatical}, they also made use of the non-public CLC corpus in training in addition to Lang-8 and NUCLE, resulting in 2.6M sentence pairs. By further adding a web-scale Common Crawl LM that was used in \cite{junczysdowmunt2016phrase} in a rescoring step, they achieved an F\textsubscript{0.5} score of 45.15 on the CoNLL-2014 test set. Their rescorer was trained using a simple grid search with fixed step size to get the feature weights and did not make use of task-specific features, whereas we use minimum error rate training \cite{och2003mert} to find optimal feature weights and use edit operation features and LM features.

More recently, \citeauthor{schmaltz2017adapting}~\shortcite{schmaltz2017adapting} used a word-level bidirectional LSTM network trained on Lang-8 and NUCLE (1.4M sentence pairs) with edit operations (insertions, deletions, and substitutions) marked with special tags in the target sentences. Their untuned model and the baseline that did not have edit operation tags marked yielded a high precision and a low recall. However, when they tuned the weights for the edit operations using a grid search maximizing F\textsubscript{0.5}, their recall went up. Without using any additional models or corpora, their approach achieved F\textsubscript{0.5} score of 41.37 on the CoNLL-2014 test set. Their edit operation tagging method and tuning also implicitly modeled edit operation weights. We model edit operations explicitly in our approach by counting and using them as weighted features in our rescorer. 

%%%%%%%%%%%%%%%%%%%%%%%%%%%%%%%%%%%%%%%%%%%%
%% Multi-Layer Convolutional Encoder-Decoder
%%%%%%%%%%%%%%%%%%%%%%%%%%%%%%%%%%%%%%%%%%%%
\section{A Multilayer Convolutional Encoder-Decoder Neural Network}

Encoder-decoder models are most widely used for machine translation from a source language to a target language. Similarly, an encoder-decoder model can be employed for GEC, where the encoder network is used to encode the potentially erroneous source sentence in vector space and a decoder network generates the corrected output sentence by using the source encoding. The attention mechanism \cite{bahdanau2015neural} selectively weights different parts of the source sentence during decoding, allowing for a different encoding of the source sentence at every decoding time step. We build our models based on an encoder-decoder architecture with multiple layers of convolutions and attention mechanisms, similar to its use in MT by \cite{gehring2017convolutional}. The models are trained on words with rare words segmented into sub-words \cite{sennrich2016neural}.

% Model
%******
\subsection{Model}

\begin{figure}[t]
  \centering
  \includegraphics[width=0.47\textwidth]{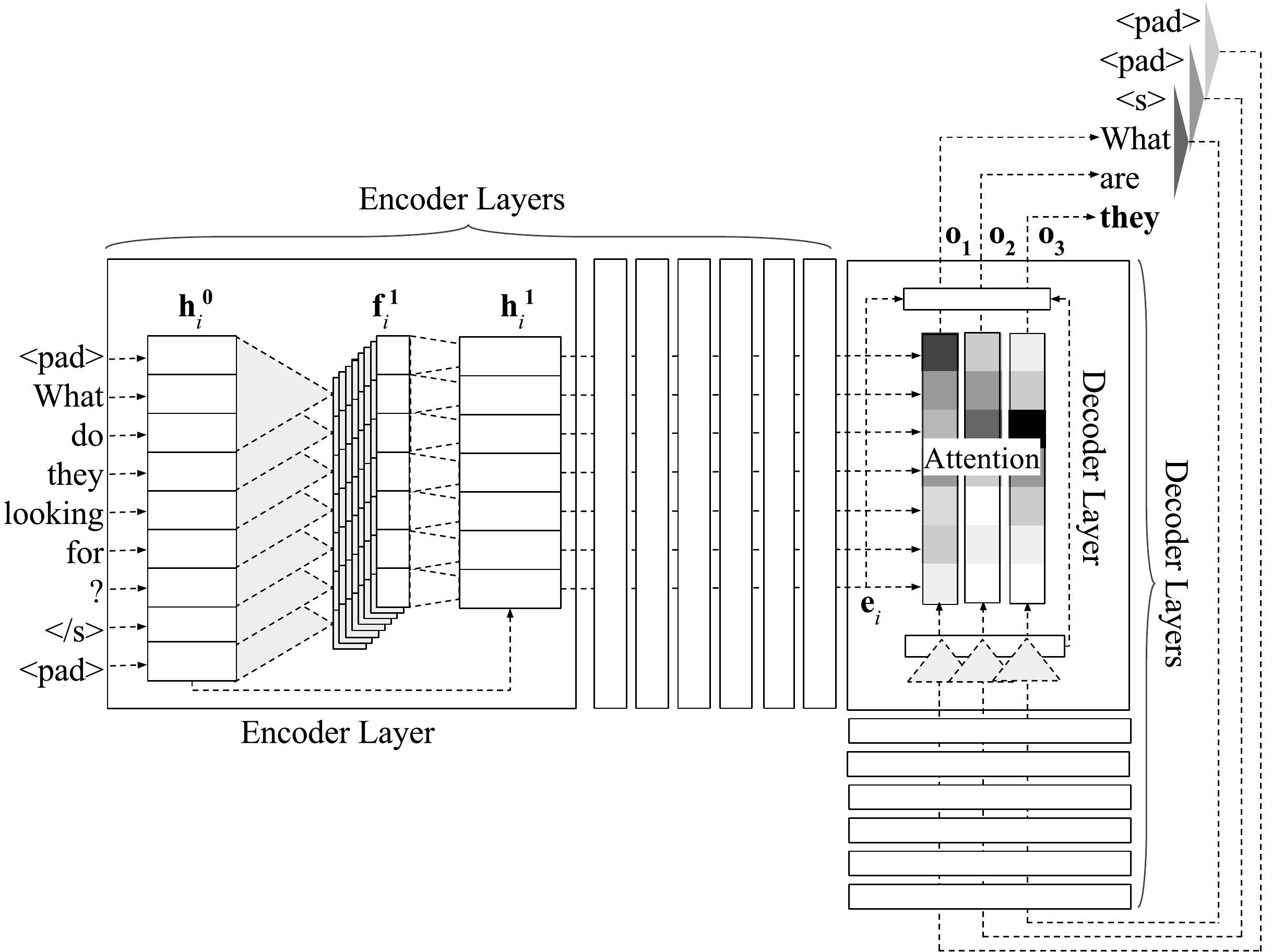}
  \caption{Architecture of our  multilayer convolutional model with seven encoder and seven decoder layers (only one encoder and one decoder layer are illustrated in detail).}
  \label{fig:architecture}
\end{figure}

Consider an input source sentence $S$ given as a sequence of $m$ source tokens $s_1,\ldots,s_m$ and $s_i \in V_s$, where $V_s$ is the source vocabulary. The last source token, $s_m$, is a special end-of-sentence marker token. The source tokens are embedded in continuous space as $\mathbf{s}_1,\ldots,\mathbf{s}_m$. The embedding $\mathbf{s}_i \in \mathbb{R}^{d}$ is given by $\mathbf{s}_i = \mathbf{w}(s_i) + \mathbf{p}(i)$, where $\mathbf{w}(s_i)$ is the word embedding and $\mathbf{p}(i)$ is the position embedding corresponding to the position $i$ of token $s_i$ in the source sentence. Both embeddings are obtained from embedding matrices that are trained along with other parameters of the network. 

The encoder and decoder are made up of $L$ layers each. The architecture of the network is shown in Figure \ref{fig:architecture}. The source token embeddings, $\mathbf{s}_1,\ldots,\mathbf{s}_m$, are linearly mapped to get input vectors of the first encoder layer, $\mathbf{h}^0_1,\ldots,\mathbf{h}^0_m$, where $\mathbf{h}^0_i \in \mathbb{R}^{h}$ and $h$ is the input and output dimension of all encoder and decoder layers. Linear mapping is done by multiplying a vector with weights $\mathbf{W} \in \mathbb{R}^{h \times d}$ and adding the biases $\mathbf{b} \in \mathbb{R}^{h}$:
\begin{align*}
\mathbf{h}^0_i = \mathbf{W} \mathbf{s}_i + \mathbf{b}
\end{align*}
In the first encoder layer, $2h$ convolutional filters of dimension $3 \times h$ map every sequence of three consecutive input vectors to a feature vector $\mathbf{f}^1_i \in \mathbb{R}^{2h}$. Paddings (denoted by \texttt{<pad>} in Figure \ref{fig:architecture}) are added at the beginning and end of the source sentence to retain the same number of output vectors as the source tokens after the convolution operations. 
\begin{align*}
\mathbf{f}^1_i = \text{Conv}(\mathbf{h}^0_{i-1}, \mathbf{h}^0_{i}, \mathbf{h}^0_{i+1})
\end{align*}
where $\text{Conv}(\cdot)$ represents the convolution operation.
This is followed by a non-linearity using gated linear units (GLU) \cite{dauphin2016language}: 
\begin{align*}
\text{GLU}(\mathbf{f}^1_i)  = \mathbf{f}^1_{i,1:h} \circ \sigma(\mathbf{f}^1_{i,h+1:2h})
\end{align*}
where $\text{GLU}(\mathbf{f}^1_i) \in \mathbb{R}^{h}$, $\circ$ and $\sigma$ represent element-wise multiplication and sigmoid activation functions, respectively, and $\mathbf{f}^1_{i,u:v}$ denotes the elements of $\mathbf{f}^1_i$ from indices $u$ to $v$ (both inclusive). The input vectors to an encoder layer are finally added as residual connections. The output vectors of the $l$\textsuperscript{th} encoder layer are given by,
\begin{align*}
\mathbf{h}^l_i = \text{GLU}(\mathbf{f}^l_i) + \mathbf{h}^{l-1}_i  && i = 1,\ldots,m
\end{align*}

Each output vector of the final encoder layer, $\mathbf{h}^L_i \in \mathbb{R}^h$, is linearly mapped to get the encoder output vector, $\mathbf{e}_i \in \mathbb{R}^d$, using weights $\mathbf{W}_\mathbf{e} \in \mathbb{R}^{d \times h}$ and biases $\mathbf{b}_\mathbf{e} \in \mathbb{R}^{d}$:
\begin{align*}
\mathbf{e}_i = \mathbf{W}_\mathbf{e}  \mathbf{h}^L_i + \mathbf{b}_\mathbf{e}
&& i = 1,\ldots,m
\end{align*}

Now, consider the generation of the target word $t_{n}$ at the $n$\textsuperscript{th} time step in decoding, with $n-1$ target words previously generated. For the decoder, paddings are added at the beginning. The two paddings, beginning-of-sentence marker and the previously generated tokens, are embedded as $\mathbf{t}_{-2},\mathbf{t}_{-1}, \mathbf{t}_0,\mathbf{t}_{1},\ldots,\mathbf{t}_{n-1}$ in the same way as source token embeddings are computed. Each embedding $\mathbf{t}_j \in \mathbb{R}^d$ is linearly mapped to $\mathbf{g}^0_{j} \in \mathbb{R}^h$ and passed as input to the first decoder layer. In each decoder layer, convolution operations followed by non-linearities are performed on the previous decoder layer's output vectors $\mathbf{g}^{l-1}_{j}$, where $j = 1,\ldots,n$:
\begin{align*}
\mathbf{y}^l_{j} = \text{GLU}(\text{Conv}(\mathbf{g}^{l-1}_{j-3},\mathbf{g}^{l-1}_{j-2},\mathbf{g}^{l-1}_{j-1})
\end{align*}
where $\text{Conv}(\cdot)$ and $\text{GLU}(\cdot)$ represent convolutions and non-linearities respectively, and $\mathbf{y}^l_j$ becomes the decoder state at the $j$\textsuperscript{th} time step in the $l$\textsuperscript{th} decoder layer. The number and size of convolution filters are the same as those in the encoder.

Each decoder layer has its own attention module. To compute attention at layer $l$ before predicting the target token at the $n$\textsuperscript{th} time step, the decoder state $\mathbf{y}^l_{n} \in \mathbb{R}^h$ is linearly mapped to a $d$-dimensional vector with weights $\mathbf{W}_\mathbf{z}\in \mathbb{R}^{d \times h}$ and biases $\mathbf{b}_\mathbf{z} \in \mathbb{R}^{d}$, adding the previous target token's embedding:
\begin{align*}
\mathbf{z}^l_{n} = \mathbf{W}_\mathbf{z}  \mathbf{y}^l_{n} + \mathbf{b}_\mathbf{z} + \mathbf{t}_{n-1}  
\end{align*}
The attention weights $\alpha ^l_{n,i}$ are computed by a dot product of the encoder output vectors $\mathbf{e}_1,\ldots,\mathbf{e}_m$ with $\mathbf{z}^l_{n}$ and normalized by a softmax:
\begin{align*}
\alpha ^l_{n,i} = \dfrac{ \text{exp} (  \mathbf{e}_i^\top \mathbf{z}^l_{n}  )  } {\sum_{k=1}^{m}  \text{exp} ( \mathbf{e}_k^\top  \mathbf{z}^l_{n}  ) } &&   i = 1,\ldots,m
\end{align*}
The source context vector $\mathbf{x}^l_{n}$ is computed by applying the attention weights to the summation of the encoder output vectors and the source embeddings. The addition of the source embeddings helps to better retain information about the source tokens.
\begin{align*}
\mathbf{x}^l_{n} = \sum_{i=1}^{m} \alpha ^l_{n,i} ( \mathbf{e}_i + \mathbf{s}_i)
\end{align*}
The context vector $\mathbf{x}^l_{n}$ is then linearly mapped to $\mathbf{c}^l_n \in \mathbb{R}^{h}$.
The output vector of the $l$\textsuperscript{th} decoder layer, $\textbf{g}^l_n$, is the summation of $\mathbf{c}^l_n$, $\mathbf{y}^l_n$, and the previous layer's output vector $\mathbf{g}^{l-1}_n$.
\begin{align*}
\mathbf{g}^l_n = \mathbf{y}^l_n  + \mathbf{c}^l_n + \mathbf{g}^{l-1}_n 
\end{align*}
The final decoder layer output vector $\mathbf{g}^L_n$ is linearly mapped to $\mathbf{d}_n \in \mathbb{R}^{d}$. Dropout \cite{srivastava2014dropout} is applied at the decoder outputs, embeddings, and before every encoder and decoder layer. The decoder output vector is then mapped to the target vocabulary size ($|V_t|$)  and softmax is computed to obtain target word probabilities.
\begin{equation*}
\mathbf{o}_n = \mathbf{W}_\mathbf{o}  \mathbf{d}_n + \mathbf{b}_\mathbf{o}   \qquad \mathbf{W}_\mathbf{o} \in \mathbb{R}^{|V_t| \times d}, \mathbf{b}_\mathbf{o} \in \mathbb{R}^{|V_t|} 
\end{equation*}
\begin{equation*}
p(t_n=w_i|t_1,\ldots,t_{n-1}, S) =  \frac{\text{exp}(o_{n,i})}{\sum_{k=1}^{|V_t|}\text{exp}(o_{n,k})}
\end{equation*}
where $w_i$ is the $i$\textsuperscript{th} word in the target vocabulary $V_t$.

% Pre-Training Embeddings
%************************
\subsection{Pre-Training of Word Embeddings}

We initialize the word embeddings for the source and target words with pre-trained word embeddings learned from a large English corpus. Rare words in this English corpus are split into BPE-based sub-word units as we use similar pre-processing for the parallel corpus that is used to train the network. The word embeddings are computed by representing a word as a bag of character \textit{N}-grams and summing the skip-gram embeddings of these character n-gram sequences, using the \textit{fastText} tool ~\cite{bojanowski2017enriching}.  These embeddings have information about the underlying morphology of words and was empirically found to perform better than initializing the network randomly or using \textit{word2vec} \cite{mikolov2013distributed} embeddings, which treat words as separate entities and have no information about the character sequences that make up the words.

% Training
%************************
\subsection{Training}

The model is trained using the negative log-likelihood loss function:
\begin{align*}
L = - \frac{1}{N}\sum_{i=1}^{N} \frac{1}{T_i}\sum_{j=1}^{T_i}\text{log}\big(p(t_{i,j}|t_{i,1},\ldots,t_{i,j-1}, S)\big)
\end{align*}
where $N$ is the number of training instances in a batch, $T_i$ is the number of tokens in the $i$\textsuperscript{th} reference sentence, $t_{i,j}$ is the $j$\textsuperscript{th} target word in the reference correction for the  $i$\textsuperscript{th} training instance. The parameters are optimized using Nesterov's Accelerated Gradient Descent (NAG) with a simplified formulation for Nesterov's momentum~\cite{bengio2013advances}. 

% Decoding
%************************
\subsection{Decoding}

The encoder-decoder model estimates the probability of target words given the erroneous source sentence $S$. The best sequence of target words is obtained by a left-to-right beam search. In a beam search, the top $b$ probable candidates at every decoding time step is retained.  The top-scoring candidate in the beam at the end of the search will be the correction hypothesis. The model score of a hypothesis is the sum of the log probabilities of the hypothesis words computed by the network. We also perform ensembling during decoding by averaging the predictions from multiple models in order to compute the log probability scores of the hypothesis words. The models used for ensembling have the same architecture but are trained with different random initializations.

% Rescoring
%************************
\subsection{Rescoring}

In order to incorporate task-specific features and large language models, we re-score the final beam candidates using a log-linear framework. The score of a correction hypothesis sentence $T$ given the source sentence $S$ is given by,
\begin{align*}
\text{score}(T,S) = \sum_{i=1}^{F}\lambda_i f_i(T,S)
\end{align*}
where, $\lambda_i$ and $f_i$ are the $i$\textsuperscript{th} feature weight and feature function respectively, and $F$ is the number of features. The feature weights are computed by minimum error rate training (MERT) \cite{och2003mert} on the development set. We use the following sets of features in rescoring in addition to the model score of the hypothesis:
\begin{enumerate}
\item \textbf{Edit operation (EO)} features: Three features denoting the number of token-level substitutions, deletions, and insertions between the source sentence and the hypothesis sentence. 
\item \textbf{Language model (LM)} features: Two features, a 5-gram language model score (i.e., the sum of log probabilities of 5-grams in the hypothesis sentence) and the number of words in the hypothesis. Similar to state-of-the-art methods, the language model is trained on the web-scale Common Crawl corpus \cite{chollampatt2017connectingdots,junczysdowmunt2016phrase}.
\end{enumerate}

%%%%%%%%%%%%%%%%%%%
%% Main Results

\begin{table*}[t]
\small
\centering
\begin{tabular}{|l|c|c|c|c|c|c|}
\hline

System	& Parallel	& Is Data & Other & \multicolumn{3}{c|}{CoNLL-2014 Test Set} \\ \cline{5-7} % \cline{8-9} 
		& Data		& Public?& Corpora& Prec. & Recall & F\textsubscript{0.5}  \\ 
\hline      
\multicolumn{7}{|l|}{\textit{Baselines}} \\ 
\hline 
SMT			& L8, NUCLE	& Yes	& --	& 57.94	& 16.48	& 38.54	\\ 
SMT +NNJM	& L8, NUCLE	& Yes	& -- 	& 58.38	& 18.83	& 41.11	\\ 
\hline
\multicolumn{7}{|l|}{\textit{This work (no additional corpora)}} \\ 
\hline
MLConv				& L8, NUCLE	& Yes	& --	& 59.68	& 23.15	& 45.36	\\ 
MLConv (4 ens.)		& L8, NUCLE	& Yes	& --	& 67.06	& 22.52	& 48.05	\\
MLConv (4 ens.) +EO	& L8, NUCLE	& Yes	& --	& 62.36	& 27.55	& 49.78	\\ 
\hline
\multicolumn{7}{|l|}{\textit{This work (using additional corpora)}} \\ 
\hline
MLConv$\textsubscript{embed}$ 						& L8, NUCLE	& Yes	& Wiki			& 60.90	& 23.74	& 46.38	\\ 
MLConv$\textsubscript{embed}$ (4 ens.)				& L8, NUCLE	& Yes	& Wiki			& 68.13	& 23.45	& 49.33	\\
MLConv$\textsubscript{embed}$ (4 ens.) +EO			& L8, NUCLE	& Yes 	& Wiki 			& 63.12	& 28.36	& 50.70	\\ 
MLConv$\textsubscript{embed}$ (4 ens.) +EO	+LM 	& L8, NUCLE	& Yes	& Wiki, CC		& 65.18	& 32.26	& 54.13	\\ 
MLConv$\textsubscript{embed}$ (4 ens.) +EO	+LM +SpellCheck	& L8, NUCLE	& Yes		& Wiki, CC, SP& 65.49 & 33.14	& 54.79	\\
\hline
\multicolumn{7}{|l|}{\textit{Prior encoder-decoder approaches}} \\ \hline 
\cite{ji2017nested}	with LM		& L8, NUCLE, CLC & No	& CC		& --	& -- 	& 45.15	\\	
\cite{ji2017nested}	without LM	& L8, NUCLE, CLC & No	& --		& --	& -- 	& 41.53	\\  
\cite{schmaltz2017adapting}		& L8, NUCLE		& Yes	& --		& --	& --	& 41.37	\\  
\cite{xie2016neural} with LM, Edit Classifier	& L8, NUCLE		& Yes	& CC (small)& 49.24	& 23.77	& 40.56	\\  
\cite{zheng2016grammatical}		& CLC			& No	& --		& --	& -- 	& 39.90	\\ 
\hline
\multicolumn{7}{|l|}{\textit{State-of-the-art systems}} \\ 
\hline 
\cite{chollampatt2017connectingdots} +SpellCheck	& L8, NUCLE	& Yes	& Wiki, CC, SP &62.74	& 32.96	& 53.14				\\ 
\cite{chollampatt2017connectingdots} 				& L8, NUCLE	& Yes	& Wiki, CC		& 62.14	& 30.92	& 51.70				\\ 
\cite{junczysdowmunt2016phrase}	& L8, NUCLE	& Yes	& Wiki, CC		& 61.27	& 27.98	& 49.49				\\
\hline 
\end{tabular}
\caption{Results on the CoNLL-2014 test set. For single models (MLConv and MLConv\textsubscript{embed}), average precision, recall, and F\textsubscript{0.5} of 4 models (trained with different random initializations) are reported. (4 ens.) refers to the ensemble decoding of these 4 models. +EO and +LM refer to re-scoring using edit operation and language model features, respectively. +SpellCheck denotes the addition of the publicly available spell checker proposed in \cite{chollampatt2017connectingdots}. L8 refers to the Lang-8 corpus, CC refers to Common Crawl, CLC refers to the non-public Cambridge Learner Corpus, and SP refers to the corpus of misspellings. A smaller subset of CC (2.2B words) was used in \cite{xie2016neural} compared to the rest (94B -- 97B words).
}
\label{tbl:main}
\end{table*}

%%%%%%%%%%%%%%%%%%%%%%
%% Setup
%%%%%%%%%%%%%%%%%%%%%%
\section{Experimental Setup}

% Datasets
%************************
\subsection{Datasets}
We use two public datasets from prior work, Lang-8 \cite{mizumoto2011mining} and NUCLE \cite{dahlmeier2013building}, to create our parallel data. Along with the sentence pairs from NUCLE, we extract and use the English sentence pairs in Lang-8 by selecting essays written by English learners and removing non-English sentences from them using a language identification tool\footnote{https://github.com/saffsd/langid.py}. Sentence pairs that are unchanged on the target side are discarded from the training set.  A subset of this data, 5.4K sentence pairs from NUCLE, is taken out to be used as the development data for model selection and training the rescorer. The remaining parallel data that is used for training the encoder-decoder NN consists of 1.3M sentence pairs (18.05M source words and 21.53M target words). We also make use of the larger English corpora from Wikipedia (1.78B words) for pre-training the word embeddings, and a subset of the Common Crawl corpus (94B words) for training the language model for rescoring. Corpora of similar size from the Common Crawl have been used by leading GEC systems \cite{chollampatt2017connectingdots,ji2017nested,junczysdowmunt2016phrase}.

% Evaluation
%************************
\subsection{Evaluation}
Our evaluation setting is the same as that in the CoNLL-2014 shared task. We evaluate our models and compare them to previous systems on the official CoNLL-2014 test set using the F\textsubscript{0.5} score computed using the MaxMatch scorer \cite{dahlmeier2012m2}. Following prior work, we analyze our neural model choices and perform ablation studies on the CoNLL-2013 shared task test set. 

We also evaluate the fluency of our model outputs on the recently released JFLEG development and test sets \cite{napoles2017jfleg}, which have fluency-based rewrites of learner-written sentences done by native writers in order to make the sentences native-sounding and error-free. The GLEU metric is used to assess fluency of corrected text when the error-span and error-type annotations are not provided \cite{napoles2015ground}. We also calculate the F\textsubscript{0.5} score after automatically extracting the annotation span using the scripts released with the JFLEG dataset.

% Model and Training Details
%***************************
\subsection{Model and Training Details}
We extend the publicly available PyTorch-based implementation\footnote{https://github.com/facebookresearch/fairseq-py} for training multilayer convolutional models initialized with pre-trained embeddings. Both source and target embeddings are of 500 dimensions. Each of the source and target vocabularies consists of 30K most frequent BPE tokens from the source and target side of the parallel data, respectively. Pre-training is done using \textit{fastText} with one pass on the Wikipedia corpus using a skip-gram model with a window size of 5. Character \textit{N}-gram sequences of size between 3 and 6 (both inclusive) are used to compute the word embeddings and other parameters are kept to their default values. The embeddings are updated during training of the encoder-decoder NN. Each of the encoder and decoder is made up of seven convolutional layers, with a convolution window width of 3. The number of layers in the encoder and decoder is set based on development set performance after experimenting with 5, 7, and 9 layers. Output of each encoder and decoder layer is of 1024 dimensions. Dropout with probability 0.2 is applied on the embeddings, convolution layers, and decoder output. We train every model simultaneously on 3 NVIDIA Titan X GPUs with a batch size of 32 on each GPU and perform validation after every epoch concurrently on another NVIDIA Titan X GPU.  A learning rate of 0.25 is used with a learning rate annealing factor of 0.1 and a momentum value of 0.99. We use early stopping and select the best model based on the F\textsubscript{0.5} score on the development set. Training a single model takes around 18 hours. During decoding, a beam width of 12 is used. 

% Baselines
%************************
\subsection{Baselines}

We compare our systems to all prior neural approaches for GEC and two state-of-the-art (SOTA) systems. The first SOTA system \cite{junczysdowmunt2016phrase} employs a word-level SMT approach with task-specific features and a web-scale LM trained on the Common Crawl corpus. The second SOTA system \cite{chollampatt2017connectingdots} adds an adapted neural network joint model (NNJM) to a word-level SMT system with task-specific features and a web-scale LM, with further improvement by spelling error correction using a character-level SMT system. In order to compare our neural approach to the SMT approach without using other English corpora, we create two baselines using released models of the SOTA system \cite{chollampatt2017connectingdots}. The first (SMT +NNJM in Table \ref{tbl:main}) is this word-level SMT-based system retuned after removing all subsidiary models that make use of additional English corpora such as the word-class LM and the web-scale Common Crawl LM. This system has an adapted NNJM and an operation sequence model (OSM), both trained on the parallel data, and has a single LM trained on the target side of the parallel data. Another non-neural SMT baseline (SMT in Table \ref{tbl:main}) is created by further removing the adapted NNJM and retuning on our development set.

%%%%%%%%%%%%%%%%%%%%%%%%%%
%% Experiments and Results
%%%%%%%%%%%%%%%%%%%%%%%%%%

\section{Experiments and Results}

% Evaluation on Benchmark Corpora
%************************
\subsection{Evaluation on Benchmark Corpora}

We evaluate our systems based on the grammaticality and fluency of their output sentences.

\subsubsection{Grammaticality}

We first evaluate different variants of our system on the CoNLL-2014 test data (Table \ref{tbl:main}). Our single model without using any additional corpora or rescoring (MLConv) achieves 45.36 F\textsubscript{0.5}. After ensembling four models (4 ens.), the performance reaches 48.05 F\textsubscript{0.5} and outperforms the previous best neural model without LM \cite{ji2017nested} (41.53 F\textsubscript{0.5}) by a large margin of 6.52 F\textsubscript{0.5}, despite the latter using much more training data including the non-public CLC. Our neural systems also substantially outperform the two comparable SMT baselines, `SMT' and `SMT +NNJM'. When re-scoring is performed with edit operation (+EO) features, the performance goes up to 49.78 F\textsubscript{0.5}, outperforming a strong SMT-based system  \cite{junczysdowmunt2016phrase} that uses task-specific features and a web-scale Common Crawl language model. Our system, on the other hand, achieves this level of performance without using any additional English corpora or pre-trained word embeddings. When we train our models by initializing with pre-trained \textit{fastText} word embeddings (MLConv\textsubscript{embed}), decode using an ensemble of four models, and rescore with edit operation features, the performance reaches 50.70 F\textsubscript{0.5}.

After adding the web-scale LM in rescoring (+LM), our approach reaches 54.13  F\textsubscript{0.5}, outperforming the best previous published result of \cite{chollampatt2017connectingdots} (F\textsubscript{0.5} = 53.14) that additionally uses a spelling correction component trained on a spelling corpus. This improvement is statistically significant ($p < 0.001$). When we make use of the spelling correction component in \cite{chollampatt2017connectingdots} (+SpellCheck), our performance reaches 54.79, a statistically significant improvement of 1.65 F\textsubscript{0.5} ($p < 0.001$) over the best previous published result, and establishes the new state of the art for English GEC. All statistical significance tests were performed using sign test with bootstrap resampling on 100 samples.

%%%%%%%%%%%%%%%%%%
% Fluency Results

\begin{table}[t]
\small
\centering
\begin{tabular}{|l|c|c|c|c|}
\hline
System									&\multicolumn{2}{c|}{JFLEG Dev}		& \multicolumn{2}{c|}{JFLEG Test}	\\ \cline{2-5}
										&F\textsubscript{0.5} & GLEU 		& F\textsubscript{0.5} & GLEU\\ 
\hline
MLConv$\textsubscript{embed}$ 			& 56.67	& 47.71	& 58.82	& 51.34	\\ 
MLConv$\textsubscript{embed}$ (4 ens.)	& 57.62	& 47.93	& 59.21	& 51.06	\\
~~~+EO									& 59.40	& 49.24 & 62.15	& 53.38	\\ 
~~~~~~+LM 								& 62.44	& 51.24	& 65.87	& 55.99	\\ 
~~~~~~~~~+SpellCheck					& 63.61	& 52.48	& 66.80	& 57.47\\
\hline
C\&N~\shortcite{chollampatt2017connectingdots} 	& 58.17	& 48.17	& 60.95	& 53.18 \\
~~~~~~~~~+SpellCheck				 	& 61.51 & 51.01	& 64.25 & 56.78 \\
\hline
\end{tabular}
\caption{Results on the JFLEG development and test sets compared to \cite{chollampatt2017connectingdots} (C\&N). For MLConv\textsubscript{embed}, average F\textsubscript{0.5} and GLEU of 4 models (trained with different random initializations) are reported.}
\label{tbl:jfleg}
\end{table}

\subsubsection{Fluency}
We also measure the fluency of the outputs on the JFLEG development and test sets (Table \ref{tbl:jfleg}). Our system with rescoring using edit operation features outperforms the state-of-the-art system with a web-scale LM without spell checking \cite{chollampatt2017connectingdots} on both datasets and metrics. This is without adding the web-scale LM to our system. After adding the web-scale LM and using the spell checker, our method achieves the best reported GLEU and F\textsubscript{0.5} scores on these datasets. It is worth noting that our models achieve this level of performance without tuning on the JFLEG development set.

\subsection{Encoder and Decoder Architecture}
\begin{table}[t]
\centering
\small
\begin{tabular}{|l|c|c|c|}
 \hline
Architecture		 		&	 Prec. & 	Recall & 	F\textsubscript{0.5} \\
\hline 
BiLSTM		&	52.49	&	10.95	&	29.84	\\ 
SLConv		&	43.65	&	10.23	&	26.39	\\ 
\hline
MLConv 		&	51.90	&	12.59	&	31.96	\\ 
\hline
 \end{tabular}
 \caption{Performance of various architectures on the CoNLL-2013 test set.}
 \label{tbl:blstmconv}
 \end{table}

We analyze the performance of various network architectures without using pre-trained word embeddings on the CoNLL-2013 test set (Table \ref{tbl:blstmconv}). We experiment with using a bidirectional LSTM in the encoder and an attentional LSTM decoder with a soft attention mechanism \cite{bahdanau2015neural} (BiLSTM in Table \ref{tbl:blstmconv}), and compare it to single layer convolutional (SLConv) as well as our proposed multilayer convolutional (MLConv) encoder and decoder models. BiLSTM can capture the entire sentence context from left and right for each input word, whereas SLConv captures only a few surrounding words (equal to the filter width of 3). However, MLConv captures a larger surrounding context ($7$ layers $\times$ filter width $3 = 21$ tokens) more effectively, causing it to outperform both SLConv and BiLSTM.

\begin{figure}[t]
\centering
  \includegraphics[width=0.47\textwidth]{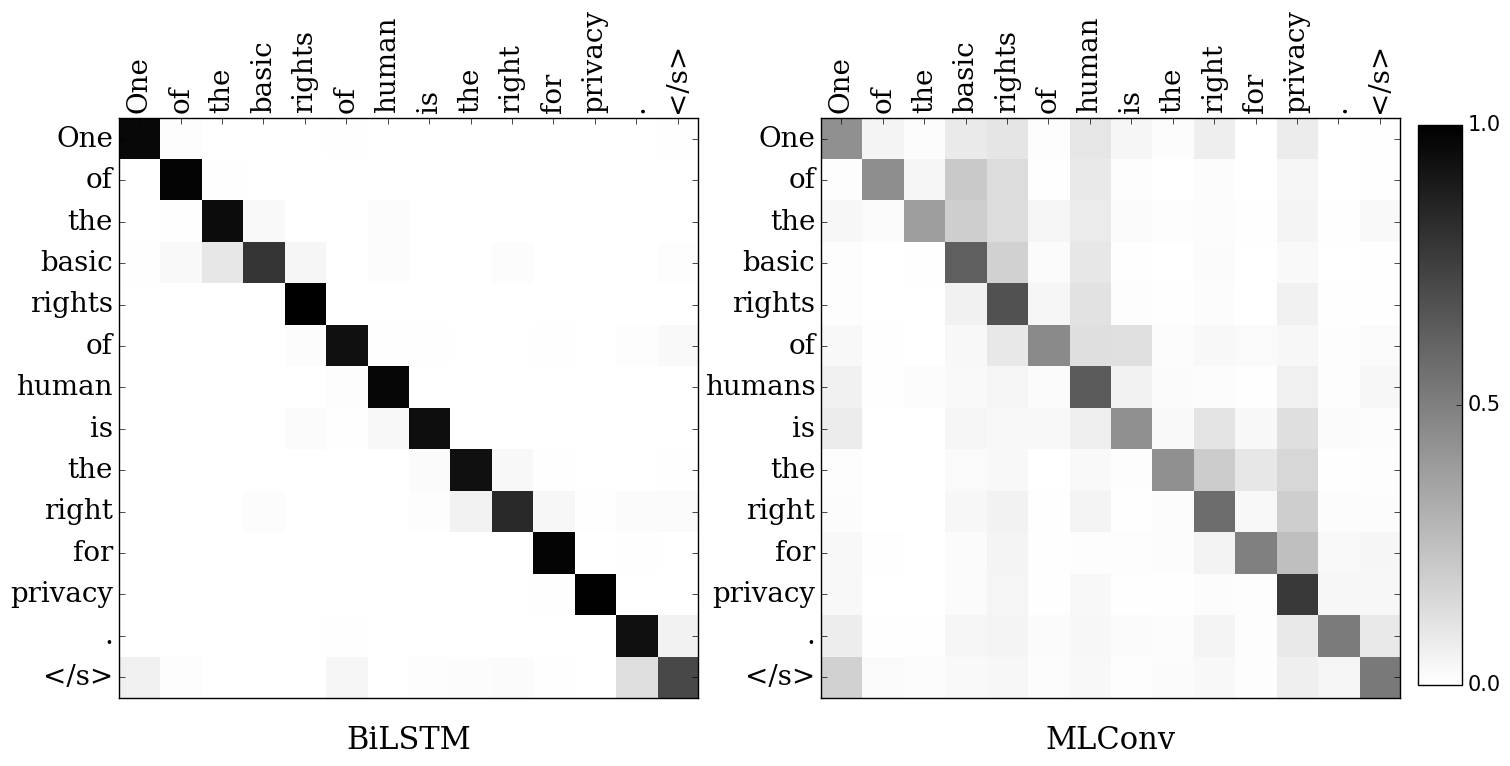}
  \caption{Visualization of attention weights of BiLSTM and MLConv models. $y$-axis shows the target words and $x$-axis shows the source words.}
  \label{fig:blstm_mlconv}
\end{figure}

It is interesting to note that the BiLSTM model has a higher precision than the MLConv model, although its recall is lower. We analyze the attention weights of both models (Figure \ref{fig:blstm_mlconv}) on an example sentence from the CoNLL-2013 test set. The attention weights shown for the MLConv model is the averaged attention weights of all decoder layers. It can be seen that BiLSTM produces a sharper distribution placing higher weights on matching source words as opposed to MLConv which places noticeable probability mass on the surrounding context words also. We observed this trend for all other examples that we tried. This could be the reason that causes BiLSTM to frequently output the source words, leading to a fewer number of proposed corrections and consequently, a higher precision. This analysis demonstrates the ability of MLConv in capturing the context better, thereby favoring more corrections than copying of the source words.

\subsection{Initialization with Pre-trained Embeddings}

We assess various methods of initializing the source and target word embeddings. Table \ref{tbl:embeddings} shows the results of initializing the embeddings randomly as well as with \textit{word2vec} and \textit{fastText} on the CoNLL-2013 test set. We train skip-gram models with \textit{word2vec} and use parameters identical to those we use for \textit{fastText}.   \textit{fastText} embeddings have access to the character sequences that make up the words and hence are better suited to learn word representations taking morphology into account. We also find that initializing with \textit{fastText} works well empirically, and hence we choose these embeddings to initialize our network when evaluating on benchmark test datasets. 

\begin{table}[t]
\centering
\small
\begin{tabular}{|l|c|c|c|}
 \hline
Initialization  &	 Prec. & 	Recall & 	F\textsubscript{0.5} 	\\
\hline 
Random 			&	51.90	&	12.59	& 31.96		\\ 
Word2vec 		&	52.80	&	12.80	& 32.49	\\ 
\hline
fastText 		&	51.08   &   13.63   & 32.97   	\\ 
\hline
\end{tabular}
\caption{Results of different embedding initializations on the CoNLL-2013 test set.}
\label{tbl:embeddings}
\end{table}

\section{Analysis and Discussion}

\begin{figure}[t]
  \includegraphics[width=0.46\textwidth]{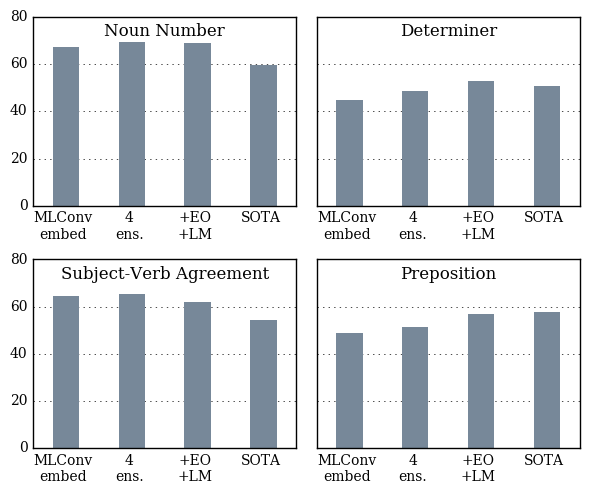}
  \caption{Performance of our models compared to the state-of-the-art system \cite{chollampatt2017connectingdots} on common error types evaluated on the CoNLL-2014 test set based on F\textsubscript{0.5}.}
  \label{fig:etypescores}
\end{figure}

We perform error type-specific performance comparison of our system and the state-of-the-art (SOTA) system \cite{chollampatt2017connectingdots}, using the recently released ERRANT toolkit \cite{bryant2017automatic} on the CoNLL-2014 test data based on F\textsubscript{0.5}. ERRANT relies on a rule-based framework to identify the error type of corrections proposed by a GEC system. The results on four common error types are shown in Figure \ref{fig:etypescores}. We find that our ensembled model with the rescorer (+EO+LM) performs competitively on preposition errors, and outperforms the SOTA system on noun-number, determiner, and subject-verb agreement errors. One of the weaknesses of SMT-based systems is in correction of subject-verb agreement errors, because a verb and its subject can be very far apart within a source sentence. On the other hand, even our single model (MLConv\textsubscript{embed}) without rescoring is superior to the SOTA SMT-based system in terms of subject-verb agreement errors, since it has access to the entire source context through the global attention mechanism and to longer target context through multiple layers of convolutions in the decoder. 

From our analysis, we find that a convolutional encoder-decoder NN captures the context more effectively compared to an RNN and achieves superior results. However, RNNs can give higher precision, so a combination of both approaches could be investigated in future. Improved language modeling has been previously shown to improve GEC performance considerably. We leave it to future work to explore the integration of web-scale LM during beam search and the fusion of neural LMs into the network. We also find that a simple preprocessing method that segments rare words into sub-words effectively deals with the rare word problem for GEC, and performs better than character-level models and complex word-character models. 

\section{Conclusion}
We use a multilayer convolutional encoder-decoder neural network for the task of grammatical error correction and achieve significant improvements in performance compared to all previous encoder-decoder neural network approaches. We utilize large English corpora to pre-train and initialize the word embeddings and to train a language model to rescore the candidate corrections. We also make use of edit operation features during rescoring. By ensembling multiple neural models and rescoring, our novel method achieves improved performance on both CoNLL-2014 and JFLEG data sets, significantly outperforming the current leading SMT-based systems. We have thus fully closed the large performance gap that previously existed between neural and statistical approaches for this task. The source code and model files used in this paper are available at https://github.com/nusnlp/mlconvgec2018.

\section*{Acknowledgements}
We thank the anonymous reviewers for their feedback. This research was supported by
Singapore Ministry of Education Academic Research
Fund Tier 2 grant MOE2013-T2-1-150.
%References and End of Paper
%These lines must be placed at the end of your paper
\bibliography{Chollampatt-Ng}

\bibliographystyle{aaai}
\end{document}